\documentclass{article}

\PassOptionsToPackage{numbers, compress}{natbib}
\usepackage[preprint]{neurips_2026}
\usepackage{enumitem}
\usepackage[utf8]{inputenc}
\usepackage[T1]{fontenc}
\usepackage{hyperref}
\hypersetup{
    colorlinks=true,
    linkcolor=red,
    citecolor=teal,
    urlcolor=cyan,
}
\usepackage{url}
\usepackage{booktabs}
\usepackage{tabularx}
\usepackage{amsfonts}
\usepackage{amsmath}
\usepackage{amssymb}
\usepackage{amsthm}
\usepackage{nicefrac}
\usepackage{microtype}
\usepackage{xcolor}
\usepackage{colortbl}
\usepackage{tikz}
\usepackage{tcolorbox}
\usepackage{multirow}
\usepackage{array}

\usetikzlibrary{arrows.meta, positioning, fit, backgrounds, calc}

\newcolumntype{Y}{>{\raggedright\arraybackslash}X}
\definecolor{hpvHighlight}{RGB}{199, 214, 161}
\definecolor{mpvHighlight}{RGB}{183, 222, 232}

\newtheorem{definition}{Definition}

\title{Stop Comparing LLM Agents \\Without Disclosing the Harness}

\author{%
\begin{tabular}{>{\raggedright\arraybackslash}p{0.25\textwidth}
                >{\raggedright\arraybackslash}p{0.25\textwidth}
                >{\raggedright\arraybackslash}p{0.25\textwidth}}
\textbf{Yunbei Zhang\textsuperscript{1}} &
\textbf{Janet Wang\textsuperscript{1}} &
\textbf{Yingqiang Ge\textsuperscript{2}} \\
\textbf{Weijie Xu\textsuperscript{3}} &
\textbf{Jihun Hamm\textsuperscript{1}} &
\textbf{Chandan K. Reddy\textsuperscript{4}} \\[5pt]
\multicolumn{3}{l}{\normalfont
\textsuperscript{1}Tulane University \quad
\textsuperscript{2}Rutgers University \quad
\textsuperscript{3}Independent Researcher \quad
\textsuperscript{4}Virginia Tech
} \\[4pt]
\multicolumn{3}{c}{\normalfont
\texttt{yzhang111@tulane.edu} \quad
\texttt{reddy@cs.vt.edu}
}
\end{tabular}
}

\begin{document}

\maketitle

\begin{abstract}
This position paper argues that, for long-horizon tasks evaluated across models with comparable frontier capability, the agent execution harness, namely the infrastructure layer that governs context construction, tool interaction, orchestration, and verification around a language model, is often a stronger determinant of agent performance than the model it wraps. We formalize and defend the \textbf{Binding Constraint Thesis}: in this regime, performance variance is governed more by harness configuration than by model choice, and current evaluation protocols therefore systematically misattribute harness-level gains to model improvements. We support this thesis along three lines. First, a control-theoretic formalization treats the harness as the controller of a closed-loop dynamical system and the LLM as the stochastic policy it governs, which explains why small harness changes can produce performance shifts that exceed those obtained by substituting one model for another. Second, published benchmarks, industry deployments, and a controlled variance decomposition show that harness-induced variance can substantially exceed model-induced variance, including cases of model ranking reversal. Third, we propose a harness-aware evaluation framework with a disclosure standard and a variance decomposition protocol. Until harness specifications are disclosed, leaderboard comparisons for long-horizon agents should be treated as incomplete and potentially misleading.
\end{abstract}

\section{Introduction}
\label{sec:intro}

The standard practice in LLM agent evaluation reports a single number per $\{\texttt{model}, \texttt{benchmark}\}$ pair and attributes it to the model. Leaderboards such as SWE-bench \cite{jimenez2023swe}, Terminal-Bench \cite{merrill2026terminal}, AgentBench \cite{liu2023agentbench}, and GAIA \cite{mialon2023gaia} treat agent performance as if it were a property of the model alone, and this convention directs research toward model scaling, fine-tuning for tool use, and prompt-level techniques. The implicit assumption is that agent capability is primarily driven by model capability, and that a sufficiently capable model produces reliable behavior on long-horizon tasks.

This assumption omits the \textit{execution harness}: the software layer between the model and the task that constructs the context the model sees, mediates its tool calls, validates its outputs, and decides when to retry, escalate, or stop. Every benchmark score is jointly produced by a model and a harness, but the harness is rarely disclosed and almost never held constant across comparisons. The phenomenon is not anecdotal: holding the model fixed while changing only the harness raises Terminal-Bench 2 pass@1 from 69.7\% to 77.0\%~\cite{lin2026agenticharnessengineeringobservabilitydriven}, and independent third-party benchmark monitoring reports up to 15 percentage points of scaffold-only variation on SWE-bench Verified~\cite{brandWhyBenchmarkingHard2025}. On such long-horizon tasks, the same model under a different harness can therefore rank above or below a competitor.

\textbf{We argue that benchmark scores for LLM agents on long-horizon tasks are not valid for cross-model comparison unless the execution harness is disclosed.} Valid comparisons require either a locked-harness protocol, where a single specified harness is applied to all models, or a factorial protocol, where harness choice is varied as a controlled factor and variance components are reported. This disclosure extends to the infrastructure layer responsible for context construction, tool-call mediation, and output verification. Without such disclosure, leaderboards measure a mixture of model and harness, and the conclusions drawn from them are systematically misattributed to the model. 

The practical consequences are concrete. Cross-paper model comparisons are unreliable, since two reported scores typically reflect different harnesses. Reproductions of a published score are difficult without a full harness specification, because the harness can be as important to the outcome as the model itself. Research incentives are also distorted: investments target model-side improvements while harness changes, the larger source of performance variance on long-horizon tasks, receive proportionally less attention. Better models still matter, and harness engineering is part of the deployed agent. The position calls for an attribution framework that decides whether an observed benchmark gain should be credited to the model, the harness, or their interaction. Beyond academic accuracy, these misattributions shape decisions the field is now making at scale, including which models to fund, which research directions to scale, and which agents to deploy. An evaluation regime that cannot separate model from harness routes those decisions to the wrong lever.

\begin{figure}[t]
\centering
\includegraphics[width=0.95\linewidth]{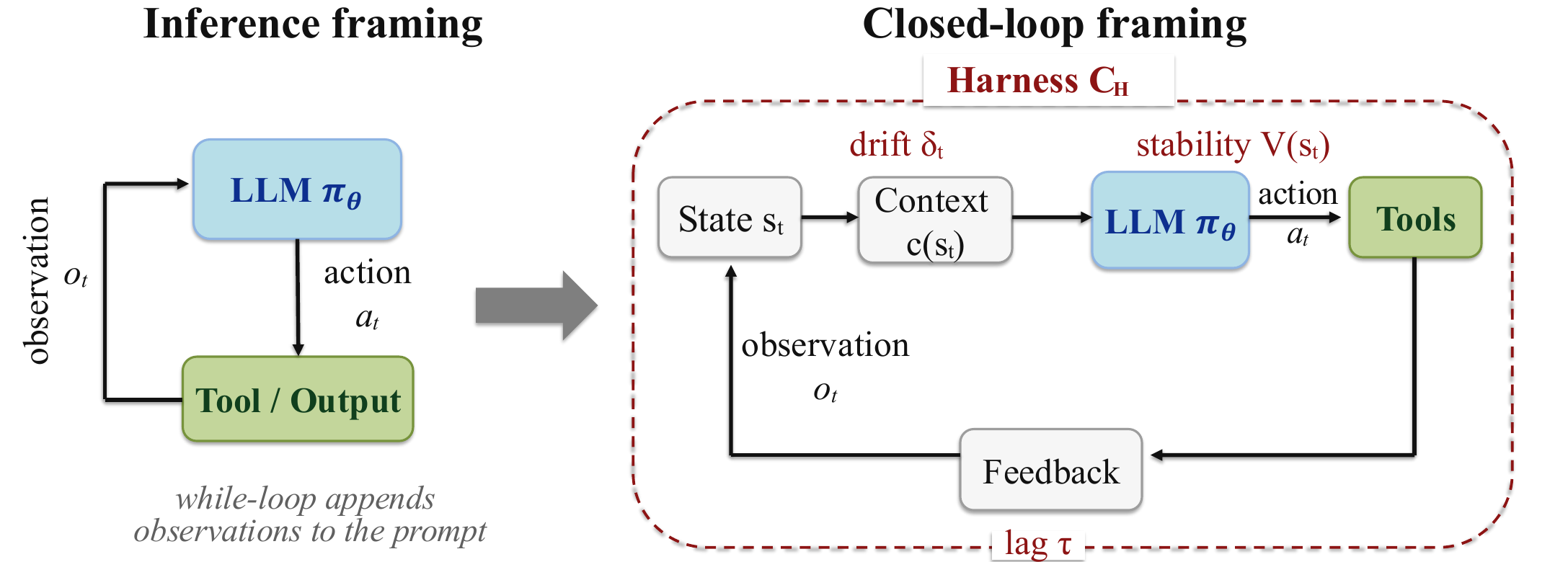}
\caption{\textbf{From inference to control.} The inference framing (left) treats the agent as a model in a while-loop, attributing performance to $\pi_\theta$. The closed-loop framing (right) makes the harness the controller $\mathcal{C}_H$: stability $V(s_t)$, context drift $\delta_t$, and control lag $\tau$ are controller properties, the model is open-loop, and the harness closes the loop.}
\label{fig:closed_loop}
\end{figure}

\section{Why Current Evaluation Conflates Model and Harness}
\label{sec:problem_formulation}

\subsection{How the Harness Enters Every Benchmark Score}
\label{sec:harness_and_score}
Standard agent evaluation queries the model with a prompt assembled from the task description, system prompt, tool catalogue, and prior trajectory state~\cite{yao2022react,yang2024sweagent,wang2024openhands}, parses the token-level output as a final answer, tool invocation, or malformed string, then routes tool calls, handles errors, compresses context as the trajectory grows, and stops after some number of steps. Every step in this pipeline (context construction, tool invocation, parsing, retry, summarization, and stopping) is a harness decision~\cite{lin2026agenticharnessengineeringobservabilitydriven,lee2026meta,lou2026autoharness,zhang2026agentic}. A single benchmark score is therefore the joint outcome of these choices and the model that generates outputs against them, but the published number records the model and elides the rest.

The conflation is not hypothetical. Under the standardized SEAL scaffold on SWE-bench Pro, Claude Opus 4.5 reaches 45.9\%, while under Claude Code the same model reaches 55.4\%. Adding a single search subagent (WarpGrep) to otherwise identical infrastructure flips the SWE-bench Pro ordering between MiniMax 2.5 and Claude Opus 4.6, despite Claude Opus ranking higher on most other benchmarks~\cite{morphSWEBenchProLeaderboard2026}. The Holistic Agent Leaderboard (HAL) reports double-digit gaps for the same model under different scaffolds on SWE-bench Verified Mini, with reported single-model swings of up to nearly 48 percentage points across leading frontier models~\cite{kapoor2025holistic}. Independent third-party benchmark monitoring reports up to 11 percentage points of scaffold-only variation for GPT-5 and 15 points for Kimi K2 Thinking on SWE-bench Verified~\cite{brandWhyBenchmarkingHard2025}. None of these effects are model upgrades. They are harness substitutions, and they routinely dwarf the 2 to 4 percentage point shifts that papers report as meaningful model advances. The resulting attribution gap matters because the same composite number is being read in three incompatible ways: as a model property in cross-paper model comparisons, as a benchmark property in tracking field progress, and as an agent property in deployment decisions. However, only the third reading is well-defined. The first two require disentanglement that current reporting does not provide.

\subsection{The Structural Reason: Agents Are Closed-Loop Systems}
\label{sec:agent_closed_loop_system}

The conflation reflects a structural fact about how agents work. Agent execution is a discrete-time closed-loop dynamical process. Let $s_t$ denote the agent's state at step $t$, comprising the full context window and any persistent memory. The model acts as a stochastic policy $a_t \sim \pi_\theta(\cdot \mid c(s_t))$, where $c$ is the context construction function implemented by the harness. The harness implements a controller $\mathcal{C}$ that updates state from $(s_t, a_t, o_t)$ to $s_{t+1}$, where $o_t$ is the environment observation, filtered and validated before being integrated into state. Different harness configurations $H$ induce different controllers $\mathcal{C}_H$ over the same model $\pi_\theta$. In this model, the LLM is open-loop. It has no direct access to $s_t$, only to $c(s_t)$, the projection of state into context that the harness chooses to expose. It retains no memory across steps beyond what $\mathcal{C}_H$ injects into $c(s_{t+1})$. It has no mechanism for self-correction beyond the feedback path that $\mathcal{C}_H$ constructs from $o_t$. Adaptation, error recovery, and long-horizon coherence are properties of $\mathcal{C}_H$, not $\pi_\theta$. This view continues a line of agent research that treats reasoning, action, and interfaces as coupled~\cite{yao2022react,li2026deepagent,pan2024training,cai2025nex,xia2025demystifying}.

This asymmetry is the structural reason for the conflation. In closed-loop systems, the controller, not the open-loop policy, governs three quantities that determine long-horizon reliability. \textbf{Stability} is whether expected progress toward a goal is non-decreasing under the controller's update rule. \textbf{Context drift} is how fast task-relevant information leaves the context window. \textbf{Control lag} is the number of steps between an anomaly being detected and a corrective signal reaching the policy. None of the three is a model property. A more capable model can lower the base rate at which anomalies arise, but the response to anomalies once they occur is a property of $\mathcal{C}_H$ that does not improve when the model is upgraded~\cite{sutton1998reinforcement}.

The lesson is broader than agents. Open-loop systems do not become reliable through bigger actuators. They become reliable through feedback control. The same structural fact applies to LLM agents on long-horizon tasks. Fig.~\ref{fig:closed_loop} illustrates the contrast: under the inference framing the model carries the reliability burden alone, while under the closed-loop framing the harness controller surrounds the policy and supplies the feedback that long-horizon execution requires.

\subsection{Why Standard Responses Fall Short}
\label{sec:standard_responses}

If the conflation is structural, the question is what to do about it. The field has converged on four classes of response. Each addresses part of the problem, but none resolves the attribution question.

\textbf{Standardization on a single harness.} Locking the harness across submissions is the most direct response, instantiated by HAL across nine benchmarks~\cite{kapoor2025holistic}, by the SWE-bench Pro unified scaffold~\cite{deng2025swe,morphSWEBenchProLeaderboard2026}, and by Epoch AI's mini-SWE-agent recommendation~\cite{brandWhyBenchmarkingHard2025}. We endorse locked-harness protocols as one of two valid evaluation regimes (Sec.~\ref{sec:thesis} names the other), but any standardized harness embeds design choices that interact with model properties and forecloses the harness-side gains current evidence shows are large. The position calls for disclosure of the harness in use, not uniformity across all evaluation. 

\textbf{Scaling the model.} A second response holds that more capable models will manage their own controllers. The empirical trend goes the other way: Claude Code redesigned rather than removed harness components as capability grew~\cite{shihiparSeeingLikeAgent2026}, the managed-agents architecture explicitly decouples model and harness because harness assumptions go stale faster than models improve~\cite{martinScalingManagedAgents2026}, and AHE produces double-digit pass-rate gains on top of frontier base models~\cite{lin2026agenticharnessengineeringobservabilitydriven}. No model upgrade compensates for the absence of feedback structure in the controller (Sec.~\ref{sec:agent_closed_loop_system}).

\textbf{Adding more benchmarks.} A third response is that with enough independent benchmarks, harness-induced noise will average out. Recent additions span freelance software work, ML engineering, contamination-resistant coding, code synthesis, visual front-end issues, and professional services~\cite{miserendino2025swe,chanmle,jainlivecodebench,zhuo2025bigcodebench,yang2025swebench,vidgen2026apex}. The bet fails: each benchmark uses its own undisclosed harness, and harness contributions do not have zero mean, so adding benchmarks adds independent $\{model, harness\}$ draws without resolving attribution within any of them. APEX-Agents finds the opposite of what the bet requires: agent failures are predominantly execution and orchestration problems rather than knowledge failures~\cite{vidgen2026apex}.

\textbf{Optimizing the harness.} A fourth response is that the field already addresses harness quality directly. ADAS searches over agentic systems~\cite{hu2024automated}, Meta-Harness and AutoHarness optimize harness code via outer-loop search~\cite{lee2026meta,lou2026autoharness}, and AHE evolves coding-agent harnesses with cross-family transfer~\cite{lin2026agenticharnessengineeringobservabilitydriven}. Agentic Context Engineering optimizes context construction as a separate surface~\cite{zhang2026agentic}, and practitioner deployment reports reach the same conclusion~\cite{rajasekaranHarnessDesignLongrunning2026,zunicBitterLessonAgent2026}. These contributions document the phenomenon this paper analyzes but answer a different question: how to make a harness better, not how to attribute observed gains. Some report results under their own harness, compounding rather than resolving attribution. Harness optimization is part of the evidence base for the position, not a substitute for it.

The structural problem requires a structural solution: disclosure. The harness used to produce a benchmark score must be part of the experimental condition, and cross-model comparisons must hold the harness fixed (locked-harness protocol) or vary it as a controlled factor (factorial protocol).

\section{The Binding Constraint Thesis}
\label{sec:thesis}

\begin{tcolorbox}[
  colback=red!2!white,
  colframe=red!40!black,
  title={\textbf{The Binding Constraint Thesis}},
  fonttitle=\bfseries,
  boxrule=0.8pt,
  arc=2pt
]
For LLM agents operating on long-horizon tasks with comparable frontier models, let $B(M, H)$ denote the benchmark score of model $M \in \mathcal{M}$ under harness $H \in \mathcal{H}$. Define:
\begin{align*}
\text{HV}(M) = \text{Var}_{H \sim P(\mathcal{H})}[B(M, H)] \qquad \text{MV}(H) = \text{Var}_{M \sim P(\mathcal{M})}[B(M, H)]
\end{align*}
The Binding Constraint Thesis asserts that, in this regime, $\text{HV}$ is often comparable to or larger than $\text{MV}$, and may dominate it in many current long-horizon agent evaluations. Current benchmark protocols report $B(M, H^*)$ for a single undisclosed $H^*$, rendering HV unmeasurable and model comparisons incomplete and potentially misleading.
\end{tcolorbox}

\subsection{Variance Decomposition}
\label{sec:variance_decomposition}

Total performance variance over a population of $\{M, H\}$ draws decomposes exactly as

\begin{equation}
\scalebox{0.95}{$
\mathrm{Var}\bigl(B(M,H)\bigr)
=
\underbrace{\mathrm{Var}_{M}\!\left[\mathbb{E}_{H}\bigl[B(M,H)\bigr]\right]}_{\mathrm{MV}}
+
\underbrace{\mathrm{Var}_{H}\!\left[\mathbb{E}_{M}\bigl[B(M,H)\bigr]\right]}_{\mathrm{HV}}
+
\underbrace{\mathrm{Var}\bigl(\text{model} \times \text{harness}\bigr)}_{\text{interaction}}.
$}
\end{equation}


The thesis asserts that the second and third terms together dominate the first in the regime considered. The interaction term is non-negligible: under the closed-loop account, a harness emphasizing self-verification helps a model with high false-confidence rates more than a model with conservative output distributions, so the same harness shift produces different gains across models. Treating model rankings as model properties requires the interaction term to be small relative to MV, a condition the thesis denies. The decomposition is the methodological core of the position. A \textit{locked-harness} protocol fixes $H = H^*$ and recovers a clean ranking of $\{B(M_i, H^*)\}$ across models, but only under that specific $H^*$. A \textit{factorial protocol} designs a grid varying both $M$ and $H$, reports $\overline{\text{HV}} = \mathbb{E}_M[\text{HV}(M)]$ and $\overline{\text{MV}} = \mathbb{E}_H[\text{MV}(H)]$ averaged across the opposite axis, the interaction term, and the count of model-pair ranking reversals across harnesses, and supports interaction analysis directly.

\subsection{Reliability Quantities of the Harness Controller}
\label{sec:reliability_quantity}

Stability, context drift, and control lag are the three structural channels through which harness configuration produces variance. Each is a property of $\mathcal{C}_H$, not of $\pi_\theta$, and each maps to an empirical failure mode that current evaluation captures only as a single composite score. Together they identify what HV is variance of: a harness shift that changes any of the three shifts $B(M, H)$ for fixed $M$.

\begin{definition}[Stability]
An agent run is stable if the Lyapunov-like measure $V(s_t) = d(s_t, \Omega^*)$ is nonincreasing in expectation under $\mathcal{C}$, where $\Omega^*$ is the set of goal-consistent states and $d$ is an appropriate task-specific distance measure. Formally: $\mathbb{E}[V(s_{t+1}) \mid s_t] \leq V(s_t)$ for all $t$. Instability manifests empirically as hallucination spirals, context overflow, and execution runaway, all of which are controller failures rather than policy failures.
\end{definition}

\begin{definition}[Context Drift]
Let $p_t$ denote the distribution of context-window embeddings at step $t$, obtained by encoding the full context window using a fixed embedding model. Context drift is $\delta_t = D_{\text{KL}}(p_t \| p_0)$, the divergence of the current context distribution from the initial task-relevant state $p_0$. A high drift rate $\dot{\delta} = \delta_t - \delta_{t-1}$ indicates that the controller is failing to maintain task-relevant information in the context window and empirically correlates with task abandonment and semantic inconsistency over long horizons. Drift is a property of the harness configuration $H$ through its context construction policy $c$. No property of the model can prevent context drift when the controller fails to manage it.
\end{definition}

\begin{definition}[Control Lag]
We propose control lag $\tau$ as a harness-specific reliability measure: the number of steps between harness anomaly detection (a tool failure, malformed output, or empty result) and the arrival of corrective action at the policy through the feedback path. Formally, if an anomaly is detected at step $t_d$ and the corrective observation $o_{t_c}$ reaches the policy at step $t_c$, then $\tau = t_c - t_d$. High $\tau$ enables failure cascades: the policy continues to act on a corrupted state for $\tau$ steps while the harness processes the error, compounding the damage before correction arrives.
\end{definition}

These three quantities sit on the structural side of the variance decomposition. Two configurations $H_1, H_2$ that differ in self-verification (affecting stability), context compression (affecting drift), or anomaly recovery (affecting control lag) each shift $B(M, H)$ for fixed $M$, contributing independently to HV. A more capable model can lower the base rate at which anomalies arise, which reduces demand on $\mathcal{C}_H$, but the response to anomalies once they occur is a property of $\mathcal{C}_H$ that does not improve when the model is upgraded. This is the structural reason the thesis is plausible. We use these quantities as conceptual tools for reasoning about $\mathcal{C}_H$, and develop their operational counterparts as trajectory-level metrics below.

\subsection{Scope, Falsifiability, and Implications}
\label{sec:scope_falsiability_implication}

\textbf{Scope.} The thesis is restricted to long-horizon tasks and comparable frontier models. Long-horizon means tasks where multiple steps of tool use, error recovery, and context management are required to reach a final outcome. The closed-loop reliability quantities only have an opportunity to act over such horizons. Comparable frontier models means a regime in which model capability gaps are not so large that MV mechanically dominates regardless of harness configuration. What can be said precisely is what falls outside the regime: short-horizon tasks and configurations where one model is dramatically more capable than the others on the relevant skills. The regime that remains is the regime where benchmark leaderboards drive research and product decisions.

\textbf{Falsifiability.} The thesis is empirical. It can be falsified by a factorial experiment that varies $\mathcal{H}$ while holding $\mathcal{M}$ fixed, separately varies $\mathcal{M}$ while holding $\mathcal{H}$ fixed, and finds $\overline{\text{MV}} > \overline{\text{HV}}$ on a long-horizon task distribution at comparable model capability. The thesis is not vacuous: such an outcome is conceivable in principle, and our controlled grid makes the variance ordering an empirical question. Concurrent psychometric work that decomposes agent ability into independent LLM and scaffold components offers an external probe in the same direction~\cite{ge2026agent}. The thesis would be in tension with that line of work if scaffold-ability variance were systematically smaller than LLM-ability variance across benchmarks, and consistent with it otherwise.

\section{Evidence}
\label{sec:evidence}

We support the thesis with two kinds of evidence: observational data from public leaderboards and a controlled factorial experiment under matched conditions.

\subsection{Evidence from Public Leaderboards}
\label{sec:evidence_leaderboards}

\textbf{SWE-bench Pro under standardized vs varied scaffolds.} On the Morph leaderboard~\cite{morphSWEBenchProLeaderboard2026}, the six leading frontier models span only 4.9 percentage points under the standardized SEAL scaffold (41.0\% to 45.9\%). Holding Claude Opus 4.5 fixed and varying only the harness widens this to 9.5 points (SEAL vs Claude Code). The within-model harness range thus exceeds the within-harness model range by roughly twofold, a lower bound on the HV / MV ratio for this slice. Adding the WarpGrep search subagent on top of identical infrastructure adds 2.1 to 2.2 points across models, comparable to a routine model upgrade and sufficient to flip the MiniMax 2.5 vs Claude Opus 4.6 ordering: a single tool addition makes the interaction term visible.

\textbf{Cross-scaffold gaps on SWE-bench Verified Mini.} HAL reports double-digit cross-scaffold gaps under SWE-Agent versus HAL Generalist, of 34 points for Claude Sonnet 4.5 (68\% to 34\%), 34 points for GPT-5 Medium (46\% to 12\%), and nearly 48 points for o4-mini~\cite{kapoor2025holistic}, much larger than the cross-model spread under either scaffold individually. The same psychometric work~\cite{ge2026agent}, in its leaderboard analysis, reaches the same qualitative conclusion under an item-response-theory model.

\textbf{Third-party benchmark monitoring.} Independent monitoring reports up to 11 to 15 percentage points of scaffold-only variation on SWE-bench Verified, with the explicit observation that scaffold choice has the single biggest impact on overall performance~\cite{brandWhyBenchmarkingHard2025}. The recommendation to use mini-SWE-agent for cross-model comparison is one realization of the locked-harness protocol. The thesis adds that the measurement gap between locked and unlocked scaffolds is what the variance decomposition makes visible.

The pattern is consistent and large. But public evidence is observational. Public harnesses differ across many dimensions and were built by different teams with different engineering budgets, so the public data do not isolate which harness components drive the variance.

\begin{table}[t]
\centering
\caption{Harness-induced performance changes on agent benchmarks (model held fixed, only infrastructure changes).}
\label{tab:harness-evidence}
\small
\begin{tabularx}{\textwidth}{@{}p{0.18\textwidth}p{0.18\textwidth}Yp{0.11\textwidth}p{0.11\textwidth}@{}}
\toprule
\textbf{Benchmark} & \textbf{Model (Fixed)} & \textbf{Harness Change} & \textbf{Layer} & \textbf{$\Delta$} \\
\midrule
SWE-bench Pro & Claude Opus 4.5 & SEAL $\to$ Claude Code & E, T, C & +9.5pp \\
 & & (45.9\% $\to$ 55.4\%) & & \\
SWE-bench Verified & Grok 4 & SWE-agent $\to$ xAI scaffold & T, C, O & +14--16pp \\
 & & (58.6\% $\to$ 72--75\%) & & \\
TerminalBench 2.0 & Fixed model & Prompt + middleware + verif. & C, S, V & +13.7pp \\
 & & (52.8\% $\to$ 66.5\%) & & \\
TerminalBench-2 & Fixed model & Automated harness opt. & All & 76.4\% \\
TerminalBench 2 & GPT-5.4 (high) & AHE harness evolution & C, T, S, O & +7.3pp \\
 & & (69.7\% $\to$ 77.0\%) & & \\
Coding benchmarks & Fixed model & Context format + tool defs & T, C & $\sim$10$\times$ \\
Agent tasks (Vercel) & Fixed model & 15 tools $\to$ 2 tools & T & 80\%$\to$100\% \\
SWE-bench Pro & Various & +WarpGrep search subagent & T & +2.1--2.2pp \\
\bottomrule
\end{tabularx}
\vspace{0.3em}

\begin{flushleft}
{\footnotesize Layers: E = Execution, T = Tool, C = Context, S = Scheduling, O = Observability, V = Verification.}

{\footnotesize Sources: SWE-bench Pro data from Morph~\cite{morphSWEBenchProLeaderboard2026}. SWE-bench Verified scaffold reports from Vals AI and Anthropic~\cite{valsaiSWEBench2026,schluntzRaisingBarSWEBench2025}. Terminal-Bench harness results from Trivedy, Meta-Harness, and AHE~\cite{trivedyImprovingDeepAgentsHarness2026,lee2026meta,lin2026agenticharnessengineeringobservabilitydriven}. Tool-reduction case from Vercel~\cite{quWeRemoved80Percent2025}. Edit-tool harness result from B{\"o}l{\"u}k~\cite{bolukHarnessProblem2026}.}
\end{flushleft}
\end{table}

\subsection{Controlled Factorial}
\label{sec:controlled_factorial}

To isolate harness-controlled variance, we vary configurations along the three reliability quantities while holding all other factors fixed, and report the resulting decomposition on a representative long-horizon coding task distribution.

\textbf{Setup.} We evaluate three frontier models, GPT-5.4~\cite{openaiIntroducingGPT542026}, Kimi K2.6~\cite{Kimi_K26MoonshotaiKimiK262026}, and GLM-5.1~\cite{GLM_51ZaiorgGLM512026}, selected because they were tightly clustered on the LLM Stats coding leaderboard,\footnote{LLM Stats coding leaderboard, \url{https://llm-stats.com/}, accessed April 23, 2026.} across three harness configurations $\{H_1, H_2, H_3\}$ on a difficulty-stratified 100-task subset of SWE-bench Verified~\cite{jimenez2023swe}. Each $(M_i, H_j)$ cell uses two independent runs with shared task order, Docker execution environment, SWE-bench evaluation pipeline, 50-step budget, and 120-second per-step timeout. Subset construction, model-selection rationale, and full harness specifications are in Appx.~\ref{app:harness_specs}.

The three harness configurations vary deliberately along the three reliability quantities:

\begin{itemize}[leftmargin=15pt, topsep=2pt, parsep=2pt]
\item $H_1$ \textbf{(Minimal)}: no context compression, verbose tool schema, no retry logic, no verification hooks, and no anomaly recovery. High drift rate, high control lag, and no stability guarantees. This is the baseline open-loop harness.
\item $H_2$ \textbf{(Improved)}: compressed context with task-relevance retrieval, minimal tool schema, and structured retry on tool failure with exponential backoff. This configuration reduces drift and provides moderate control lag with basic closed-loop feedback.
\item $H_3$ \textbf{(Full)}: $H_2$ plus per-step self-checking, KL-style drift checks every five steps, anomaly-detection middleware for repeated action loops and context contradictions, full output validation, and checkpoint rollback retaining the last 10 states. This configuration has low drift, low control lag, and explicit stability enforcement.
\end{itemize}

\textbf{Metrics.} We report pass@1 $B(M_i, H_j)$, model-induced variance per harness $\text{MV}(H_j) = \text{Var}_M[B(M, H_j)]$, harness-induced variance per model $\text{HV}(M_i) = \text{Var}_H[B(M_i, H)]$, the aggregate ratio $\overline{\text{HV}}/\overline{\text{MV}}$ averaged across the opposite axis, and the count of model-pair ranking reversals across harnesses.

\begin{table}[t]
\centering
\caption{Benchmark scores $B(M, H)$ on SWE-bench Verified subset100. Cells report mean pass@1 percentages over two runs, with percent signs omitted. Run-level results are reported in Appx.~\ref{app:harness_specs}, Table~\ref{tab:run-level-factorial}. The HV column and MV row report variances in pp$^2$.}
\label{tab:factorial}
\small
\begin{tabular*}{\textwidth}{@{\extracolsep{\fill}}lccc|>{\columncolor{hpvHighlight}}c@{}}
\toprule
 & $H_1$ (Minimal) & $H_2$ (Improved) & $H_3$ (Full) & \textbf{HV$(M)$} \\
\midrule
GLM-5.1 & 52.5 & 56.5 & 65.5 & \textbf{29.56} \\
GPT-5.4 & 55.0 & 58.5 & 63.5 & \textbf{12.17} \\
Kimi K2.6 & 52.0 & 59.0 & 60.5 & \textbf{13.72} \\
\midrule
\rowcolor{mpvHighlight}\textbf{MV$(H)$} & \textbf{1.72} & \textbf{1.17} & \textbf{4.22} & \cellcolor{white} \\
\bottomrule
\end{tabular*}
\vspace{0.2em}

{\footnotesize Aggregate $\overline{\text{HV}}/\overline{\text{MV}}$ ratio: 7.80$\times$. \quad Ranking reversal pairs: 6 out of 9 model-pair/harness-pair comparisons.}
\end{table}

\textbf{Findings.} The grid is consistent with the Binding Constraint Thesis on this task distribution. Average HV is $18.48$~pp$^2$ versus average MV of $2.37$~pp$^2$, a ratio of $7.80\times$. Changing the harness moves GLM-5.1 by 13.0 percentage points and GPT-5.4 and Kimi K2.6 by 8.5 points each. Changing the model within a fixed harness moves scores by only 3.0, 2.5, and 5.0 points for $H_1$, $H_2$, $H_3$. The interaction term is visible as six ranking reversals across the nine possible model-pair / harness-pair comparisons. We do not claim that the $7.80\times$ ratio is universal. The grid demonstrates that harness variance can dominate model variance under controlled conditions and that the decomposition protocol produces interpretable estimates on a realistic task distribution.

Trajectory-log analysis confirms that the $H_1 \to H_2$ shift reduces control noise rather than increasing model knowledge, while the $H_2 \to H_3$ gain comes primarily from closing the verification and recovery loop. The per-paragraph mechanism analysis and the cross-model variance pattern across $H_1$, $H_2$, $H_3$ are in Appx.~\ref{app:mechanism}.

\section{A Harness-Aware Evaluation Framework}
\label{sec:framework}

We argue that long-horizon agent evaluation should adopt a harness-aware framework with three components: a structured disclosure card, a variance decomposition protocol, and a short set of trajectory-level metrics that together make harness variance reportable, interpretable, and attributable.

\textbf{The harness card.} Every agent benchmark submission should include a \emph{Harness Card}: a structured disclosure of the harness configuration organized by the seven-layer ETCSOVG taxonomy, intended as an attention checklist rather than a design constraint. The seven layers cover \textbf{E}xecution (runtime substrate, sandboxing, step and task budgets), \textbf{T}ool (tool list, schemas, error contract), \textbf{C}ontext (window cap, compression and retrieval policy, persistent memory), \textbf{S}cheduling (agent loop, retry and escalation rules), \textbf{O}bservability (logged artifacts and traces), \textbf{V}erification (validation, self-checking, anomaly detection), and \textbf{G}overnance (permission model, side-effect boundaries, human approval points). Field-level disclosure requirements per layer are in Appx.~\ref{app:etcsovg} (Table~\ref{tab:etcsovg-fields}). The goal is to make configuration visible, not to standardize it, so that a reader comparing two cards can locate whether a score difference is attributable to the model, the harness, or their interaction.

\textbf{Variance decomposition protocol.} The variance decomposition becomes a protocol when an evaluation runs the same model under at least two meaningfully different harnesses and the same harness under at least two models on a fixed task set. The minimal valid design is a $2 \times 2$ model-by-harness grid with task order, execution environment, evaluation script, API parameters, and stopping rules held constant. A harness difference counts as meaningful only if it changes at least one ETCSOVG layer in a way expected to affect stability, drift, or control lag, for example retrieval-based context compression, a change to the tool schema and error format, or added verification and recovery hooks. For a designed $\{M_i\} \times \{H_j\}$ grid, we recommend reporting four statistics alongside the headline score: HV per model and MV per harness, the aggregate ratio $\overline{\text{HV}} / \overline{\text{MV}}$, the count of model-pair ranking reversals across harnesses, and the partial eta-squared coefficient
\begin{equation}
\scalebox{1.1}{$
\eta^2_p = 
\frac{SS_{\mathrm{interaction}}}
{SS_{\mathrm{interaction}} + SS_{\mathrm{error}}}
$}
\end{equation}
where $SS$ denotes the sum of squares from a fixed-effects two-way ANOVA. Three caveats apply: $\eta^2_p$ is a fixed-effects quantity on the chosen grid (if harnesses are sampled from a larger population the appropriate quantity is the variance component $\sigma^2_{MH}$ from a mixed-effects model), it is positively biased in small grids and should be reported alongside $\omega^2$ or a bootstrap interval, and a large $\eta^2_p$ is neither necessary nor sufficient for ranking flips, so it should be paired with explicit reversal counts.

\textbf{Trajectory-level metrics.} Aggregate pass rates leave open which controller-side mechanism is producing HV when it is large. We propose three trajectory-level metrics that operationalize the three reliability quantities and let HV shifts be associated with specific harness layers. We do not estimate these metrics on the $3 \times 3$ grid. Instead, we specify them as reporting requirements for future harness-aware evaluations. \emph{Recovery Rate $RR(k)$} operationalizes Stability as the probability that a trajectory transitions from a detected anomaly state (tool error, malformed output, failed validation, rejected patch) back to a task-advancing state within $k$ steps. Reporting $RR$ as a curve over $k \in \{1, 3, 5, 10\}$ is preferable to a single number because the choice of $k$ trades promptness against eventual recovery. \emph{Context Retention} operationalizes Context Drift as the fraction of task-relevant files, tests, and constraints present in the constructed context at each step, an auditable proxy that is monotone with respect to $\delta_t$. \emph{Control Lag $\tau$} is measured directly: the number of steps between a harness-detected anomaly and the arrival of a corrective signal at the policy. Together with the variance decomposition, these metrics support attribution rather than just measurement: a shift in $RR$ implicates verification and recovery layers, a shift in $\tau$ implicates observability and tool layers, and a shift in Context Retention implicates the context construction policy. The resulting attribution is descriptive co-movement, not causal. Layer-ablation designs that toggle a single ETCSOVG layer at a time are the natural extension once these metrics are routinely collected. Two complementary diagnostics extend the framework as future work for the community: \emph{Schema Compliance Rate} measures the fraction of model outputs that parse without harness intervention (interface alignment), and \emph{Action Efficiency} measures the ratio of productive, task-advancing actions to total actions (trajectory productivity). Both supplement, rather than substitute for, the three reliability quantities.
\section{Alternative Views and Counterarguments}
\label{sec:counterarguments}

Four counterarguments deserve a direct response: that scaling models will dissolve the problem, that standardization is the right reply, that interaction effects are small enough to ignore, and that the harness, as part of the deployed agent, cannot be cleanly separated in evaluation.

\textbf{Model capability will dissolve the harness problem.} One might argue that sufficiently capable models will manage their own context and recovery, removing the need for external infrastructure. The empirical trend goes the other way: the harness ecosystem has grown more complex as models have improved, with Claude Code redesigning rather than removing harness components, the managed-agents architecture explicitly decoupling model and harness, and AHE producing double-digit gains on top of frontier base models~\cite{shihiparSeeingLikeAgent2026,martinScalingManagedAgents2026,lin2026agenticharnessengineeringobservabilitydriven}. The closed-loop account explains why: long-horizon reliability is a controller property, and no model upgrade compensates for the absence of feedback structure.

\textbf{Standardization is enough.} One might concede that harness variance is large but argue for standardizing the harness across the field rather than rethinking reporting. We partially endorse it: locked-harness protocols are one of the two valid regimes. But any standardized harness embeds design choices that interact with model properties, as Epoch AI itself flags when recommending mini-SWE-agent~\cite{brandWhyBenchmarkingHard2025}, and locking the harness forecloses harness-side gains that current evidence shows are large~\cite{lin2026agenticharnessengineeringobservabilitydriven,lee2026meta,lou2026autoharness}. Standardization is also unlikely to be uniform in practice: agent harnesses manage stateful execution across multiple layers, and vendors have strong competitive incentives to differentiate at the runtime layer. Whether the harness is locked or varied, the position is that the configuration must be disclosed and treated as part of the experimental condition.

\textbf{Interaction effects are small enough to ignore.} One might concede that HV is large but argue that the interaction term is small, so any one harness still recovers a stable ordering. SWE-bench Pro data refute this: under SEAL the top six frontier models span only 4.9 points, and adding the WarpGrep subagent flips MiniMax 2.5 versus Claude Opus 4.6~\cite{morphSWEBenchProLeaderboard2026}. The LangChain Terminal-Bench result shows the same coupling, where a harness tuned around one model yields a lower score when paired with another~\cite{trivedyImprovingDeepAgentsHarness2026}. Our controlled grid confirms this under matched conditions: six ranking reversals across nine model-pair / harness-pair comparisons.

\textbf{The harness is part of the deployed agent, so separating is artificial.} One might argue that since deployed agents are (model, harness) compositions, attributing performance to one factor is a category error. This argument is correct as a description of deployment, but deployment optimizes the agent as a whole while \emph{evaluation that compares models} must isolate the model contribution. Harness improvements are real gains that deployments rightly capture. The position is that benchmark scores read as model rankings must disclose the harness, because otherwise the rankings are not informative about the quantity readers are trying to measure. The disclosure standard and variance decomposition protocol keep deployment-style and model-comparison-style attribution from being conflated.

\section{Conclusion and Discussion}
\label{sec:discussion}

The position has three audiences. For \emph{researchers}, reviewers should ask ``what harness was used?'' as routinely as they ask about hyperparameters and decoding settings, since a model score without a harness specification is missing part of the experimental condition. For \emph{benchmark designers}, agent benchmarks should expose harness variation as a first-class evaluation dimension, either through locked-harness tracks for clean model comparison or factorial tracks that report variance decompositions, building on programs underway in HAL~\cite{kapoor2025holistic}, the unified scaffold of SWE-bench Pro~\cite{deng2025swe}, and standardization recommendations from Epoch AI~\cite{brandWhyBenchmarkingHard2025}. For \emph{practitioners}, model selection alone is an incomplete optimization loop: in our controlled grid, moving from $H_1$ to $H_3$ shifts pass@1 by 8.5 to 13.0 percentage points at a fixed model, while changing the model at a fixed harness shifts pass@1 by 2.5 to 5.0 points. The engineering object is not only the model endpoint but the context, tool, recovery, verification, and governance surfaces through which the model operates.

\textbf{Open questions.} The position raises four questions that remain open for the community. First, the operational definition of comparable frontier models is currently informal and anchored to leaderboard proximity that is itself harness-confounded, so a non-confounded test for comparability is a prerequisite for cross-paper meta-analysis of HV / MV ratios. Second, harness diversity needs a principled notion of distance between configurations, without which $\text{Var}_H[B(M, H)]$ depends on the sampling distribution $P(\mathcal{H})$ in ways not yet standardized. Third, the right balance between disclosure and locked-harness evaluation is a community decision rather than a purely technical one, since disclosure alone may not address adversarial harness selection while locking the harness slows infrastructure innovation. Fourth, the trajectory-level diagnostics admit natural extensions for the community to develop: \emph{Schema Compliance Rate} and \emph{Action Efficiency} would supplement the three reliability quantities as interface-alignment and productivity probes, and the perturbation stress-test detailed in Appx.~\ref{app:perturbation} would test whether controller-side mechanisms reduce output instability under matched perturbations.

The execution harness is the binding constraint on long-horizon LLM-agent performance in the regime where benchmark leaderboards drive research and product decisions. The model is an open-loop policy and the harness is the controller that closes the loop, so reliability over a horizon is a controller property. Long-horizon agent evaluation should disclose Harness Cards, report variance decompositions when feasible, and include trajectory-level metrics that make recovery, drift, and control lag visible. Until those practices are routine, leaderboard comparisons for long-horizon agents should be treated as incomplete and potentially misleading.

\clearpage
\bibliographystyle{plainnat}
\bibliography{reference}

\clearpage
\appendix

\section{Example ETCSOVG Disclosure Card}
\label{app:etcsovg}

Table~\ref{tab:etcsovg-fields} gives the full field set that we expect a benchmark submission to disclose. The compact example in Table~\ref{tab:etcsovg-card-example} instantiates these fields for $H_3$ in our controlled experiment.

\begin{table}[h]
\centering
\caption{ETCSOVG disclosure fields. A submission can provide more detail, but omitting any of these fields makes it difficult to separate model effects from harness effects.}
\label{tab:etcsovg-fields}
\small
\begin{tabularx}{\textwidth}{@{}p{0.16\textwidth}Y@{}}
\toprule
\textbf{Layer} & \textbf{Minimum disclosure} \\
\midrule
Execution & Runtime substrate, sandboxing, filesystem access, network access, task timeout, step timeout, maximum steps, and evaluation entry point. \\
Tool & Tool list, schema style, tool-selection policy, error format, retryable error classes, and whether tools are model-visible or hidden middleware. \\
Context & Context cap, ordering policy, summarization/compression policy, retrieval method, persistent memory, and cache policy. \\
Scheduling & Agent loop, stopping rule, retry policy, escalation behavior, delegation policy, and rollback behavior. \\
Observability & Logged artifacts, trajectory format, checkpoints, traces, validation logs, and whether failures can be audited after the run. \\
Verification & Output parser, schema validation, self-checking, test execution, anomaly detection, and final patch validation. \\
Governance & Permission model, allowlist/denylist rules, side-effect boundaries, secret handling, and human approval points. \\
\bottomrule
\end{tabularx}
\end{table}

\begin{table}[h]
\centering
\caption{Compact ETCSOVG disclosure example for the $H_3$ configuration in our controlled experiment.}
\label{tab:etcsovg-card-example}
\small
\begin{tabularx}{\textwidth}{@{}p{0.15\textwidth}p{0.22\textwidth}Y@{}}
\toprule
\textbf{Layer} & \textbf{Disclosure field} & \textbf{$H_3$ setting} \\
\midrule
Execution & Runtime and limits & Docker SWE-bench runner; 50 steps; 120s step timeout \\
Tool & Interface and errors & Task-relevant tools; minimal schema; structured errors \\
Context & Memory policy & 32k token cap; summarize old steps; BM25 top-5 retrieval \\
Scheduling & Retry and recovery & 3-attempt backoff; checkpoint rollback depth 3 \\
Observability & Trace surface & Full trace logs; 10 retained checkpoints \\
Verification & Validation hooks & Self-checking; full output validation; anomaly detection \\
Governance & Permission model & Allowlist-oriented tool governance \\
\bottomrule
\end{tabularx}
\end{table}

\section{Harness Configuration Specifications}
\label{app:harness_specs}

\textbf{Model selection.} The three models were selected before running the grid because they were tightly clustered on the LLM Stats coding leaderboard at selection time: on the April~23, 2026 snapshot, Kimi K2.6, GLM-5.1, and GPT-5.4 had coding scores of 45.4, 45.3, and 44.6, respectively.\footnote{\url{https://llm-stats.com/leaderboards/best-ai-for-coding}, accessed April~23, 2026.} The live leaderboard updates continuously as arena votes and benchmark columns change, so current readings may differ; the snapshot is used only to justify that the models were comparable when selected.

\textbf{Subset construction.} The controlled experiment uses SWE-bench Verified test tasks sampled into ``subset100'' with seed 42 and stratification by the SWE-bench difficulty label, preserving the approximate difficulty mix of the 500-task split (39 tasks $<15$~min, 52 tasks 15~min--1~hour, 8 tasks 1--4~hours, 1 task $>4$~hours). Each model-harness cell submits all 100 tasks in each of two final runs. Table~\ref{tab:subset100} reports the subset composition; the ordered instance IDs are part of the experiment artifact.

\begin{table}[h]
\centering
\caption{SWE-bench Verified subset100 construction.}
\label{tab:subset100}
\small
\begin{tabular}{@{}lr@{}}
\toprule
\textbf{Field} & \textbf{Value} \\
\midrule
Source dataset & princeton-nlp/SWE-bench\_Verified \\
Split & test \\
Full split size & 500 instances \\
Subset size & 100 instances \\
Sampling rule & stratified by difficulty \\
Random seed & 42 \\
$<15$ min fix & 39 \\
15 min--1 hour & 52 \\
1--4 hours & 8 \\
$>4$ hours & 1 \\
\bottomrule
\end{tabular}
\end{table}

Table~\ref{tab:model-settings} reports the model-side settings held fixed across harnesses. The experiment is intentionally a harness study rather than a decoding study: all three models are called through their official API services in their default configurations, preserving the provider-supplied model behavior as closely as possible. Each model uses the same endpoint configuration in $H_1$, $H_2$, and $H_3$, and no model receives harness-specific API tuning~\cite{openaiIntroducingGPT542026,Kimi_K26MoonshotaiKimiK262026,GLM_51ZaiorgGLM512026}.

\begin{table}[h]
\centering
\caption{Model API settings used in the controlled grid.}
\label{tab:model-settings}
\small
\begin{tabularx}{\textwidth}{@{}p{0.16\textwidth}p{0.20\textwidth}Yp{0.12\textwidth}@{}}
\toprule
\textbf{Model} & \textbf{Provider} & \textbf{Generation settings} & \textbf{Timeout} \\
\midrule
GPT-5.4 & OpenAI & max output 4096 & 180s \\
Kimi K2.6 & Moonshot & max output 4096 & 180s \\
GLM-5.1 & ZAI & max output 4096 & 180s \\
\bottomrule
\end{tabularx}
\end{table}

Before averaging the two final runs in Table~\ref{tab:factorial}, we inspect each run separately. Table~\ref{tab:run-level-factorial} reports the two run-level $3 \times 3$ grids and recomputes HV and MV using the same population-variance convention as the main table. The first run gives an HV/MV ratio of 8.72$\times$, and the second gives 6.76$\times$, so the harness-dominance pattern is not an artifact of averaging.

\begin{table}[t]
\centering
\caption{Run-level pass@1 scores and variance decomposition before averaging across the two final runs. Score cells are pass@1 percentages on 100 tasks, with percent signs omitted. The highlighted HV column and MV rows are population variances over percentage-point scores, in pp$^2$.}
\label{tab:run-level-factorial}
\small
\begin{tabular*}{\textwidth}{@{\extracolsep{\fill}}lccc|>{\columncolor{hpvHighlight}}c@{}}
\toprule
 & $H_1$ (Minimal) & $H_2$ (Improved) & $H_3$ (Full) & \textbf{HV$(M)$} \\
\midrule
\multicolumn{5}{@{}l}{\textbf{First final run}} \\
GLM-5.1 & 52.0 & 57.0 & 66.0 & \textbf{33.56} \\
GPT-5.4 & 55.0 & 59.0 & 64.0 & \textbf{13.56} \\
Kimi K2.6 & 52.0 & 59.0 & 61.0 & \textbf{14.89} \\
\midrule
\rowcolor{mpvHighlight}\textbf{MV$(H)$} & \textbf{2.00} & \textbf{0.89} & \textbf{4.22} & \cellcolor{white} \\
\midrule
\multicolumn{5}{@{}l}{\textbf{Second final run}} \\
GLM-5.1 & 53.0 & 56.0 & 65.0 & \textbf{26.00} \\
GPT-5.4 & 55.0 & 58.0 & 63.0 & \textbf{10.89} \\
Kimi K2.6 & 52.0 & 59.0 & 60.0 & \textbf{12.67} \\
\midrule
\rowcolor{mpvHighlight}\textbf{MV$(H)$} & \textbf{1.56} & \textbf{1.56} & \textbf{4.22} & \cellcolor{white} \\
\bottomrule
\end{tabular*}
\vspace{0.2em}

{\footnotesize First run: mean HV = 20.67, mean MV = 2.37, HV/MV = 8.72$\times$. Second run: mean HV = 16.52, mean MV = 2.44, HV/MV = 6.76$\times$.}
\end{table}

Table~\ref{tab:harness-specs} expands the compact ETCSOVG card into the actual experimental specification. The first rows list shared controls that were intentionally held fixed; the remaining rows enumerate every harness mechanism varied in the final configurations. This is the level of detail needed to reproduce the variance decomposition and to determine whether a future comparison changes the model, the harness, or both.

\begin{table}[t]
\centering
\caption{Full harness specification for the controlled experiment. The table includes shared runtime controls and harness mechanisms varied across $H_1$, $H_2$, and $H_3$.}
\label{tab:harness-specs}
\scriptsize
\begin{tabularx}{\textwidth}{@{}p{0.16\textwidth}YYY@{}}
\toprule
\textbf{Field} & \textbf{$H_1$ Minimal} & \textbf{$H_2$ Improved} & \textbf{$H_3$ Full} \\
\midrule
Source tasks & SWE-bench Verified subset100, seed 42, fixed task order & Same as $H_1$ & Same as $H_1$ \\
Runtime & Docker SWE-bench execution and evaluation pipeline & Same as $H_1$ & Same as $H_1$ \\
Step budget & 50 agent steps; 120s per-step timeout & Same as $H_1$ & Same as $H_1$ \\
Design goal & Open-loop baseline with minimal intervention & Tool-robust closed-loop harness & $H_2$ plus self-checking and recovery controls \\
Context strategy & Append all prior steps chronologically & Compressed retrieval context & Same as $H_2$ \\
Context cap & 200k token estimate & 32k token estimate & 32k token estimate \\
History compression & None & Summarize older steps outside the recent window; compression window 8 & Same as $H_2$ \\
Retrieval & None & BM25 top-5 over older trajectory steps & Same as $H_2$ \\
Context ordering & Chronological & Relevance then recency & Relevance then recency \\
Drift monitoring & Disabled & Disabled & KL-style drift check every 5 steps; threshold 2.0 \\
Tool schema & Verbose tool descriptions & Minimal task-focused tool schema & Minimal task-focused tool schema \\
Tool exposure & All available tools & Task-relevant tool subset & Task-relevant tool subset \\
Tool error format & Raw tool errors & Structured error feedback & Structured error feedback \\
Runtime retry & No retry on tool failure & 3-attempt exponential backoff & $H_2$ retry policy plus malformed-output retry \\
Retryable errors & None & timeout, rate limit, transient network & timeout, rate limit, transient network, malformed output \\
Failure escalation & No explicit recovery policy & Continue after recoverable tool failures and feed the error back as an observation & Checkpoint-style recovery on detected anomalies \\
Parser behavior & Basic parser with one tolerated malformed-output retry & Parser normalization for natural model outputs & Parser normalization plus validation layer \\
Output validation & Disabled & Schema-only validation & Full output validation \\
Empty patch handling & Allowed & Rejected when validation is active & Rejected when validation is active \\
Self-verification & Disabled & Disabled & Enabled; lightweight per-step self-check \\
Anomaly detection & Disabled & Disabled & Enabled for repeated action loops and context contradictions \\
Checkpointing & Disabled & Disabled & Enabled; retain last 10 state checkpoints \\
Rollback policy & None & None & Roll back up to 3 checkpoints on handled anomalies \\
Observability & Minimal logs & Full trajectory-level logs & Full trace logs with verification metadata \\
Governance & Permissive permission model & Permissive permission model & Allowlist-oriented tool governance \\
\bottomrule
\end{tabularx}
\end{table}

Resource accounting complements the pass-rate results. Table~\ref{tab:resource-summary} aggregates the selected final artifacts for all three models and all three harnesses, reported as if the runs had been executed sequentially. Sequential time is an accounting quantity, not the wall-clock time of a parallel campaign. In aggregate, $H_2$ is the cheapest controller in this implementation, while $H_3$ spends more on verification and recovery to obtain the best pass rate; per-model costs still vary with provider pricing and tokenization.

\begin{table}[h]
\centering
\caption{Resource accounting for the selected final artifacts, grouped by model. Sequential time sums per-cell elapsed time and does not assume parallel execution.}
\label{tab:resource-summary}
\small
\begin{tabular*}{\textwidth}{@{\extracolsep{\fill}}llrrr@{}}
\toprule
\textbf{Model} & \textbf{Harness} & \textbf{Cost (USD)} & \textbf{Tokens} & \textbf{Sequential time} \\
\midrule
\multicolumn{5}{@{}l}{\textbf{GPT-5.4}} \\
 & $H_1$ Minimal & \$34.19 & 18.9M & 3.6h \\
 & $H_2$ Improved & \$22.19 & 9.7M & 2.0h \\
 & $H_3$ Full & \$23.05 & 10.0M & 2.4h \\
 & Subtotal & \$79.43 & 38.5M & 8.0h \\
\midrule
\multicolumn{5}{@{}l}{\textbf{Kimi K2.6}} \\
 & $H_1$ Minimal & \$24.29 & 60.4M & 19.6h \\
 & $H_2$ Improved & \$30.45 & 43.7M & 11.0h \\
 & $H_3$ Full & \$33.83 & 47.7M & 20.7h \\
 & Subtotal & \$88.57 & 151.8M & 51.3h \\
\midrule
\multicolumn{5}{@{}l}{\textbf{GLM-5.1}} \\
 & $H_1$ Minimal & \$12.54 & 29.7M & 9.2h \\
 & $H_2$ Improved & \$13.86 & 16.5M & 7.1h \\
 & $H_3$ Full & \$21.39 & 24.5M & 8.8h \\
 & Subtotal & \$47.79 & 70.7M & 25.1h \\
\midrule
\textbf{Total} & & \textbf{\$215.79} & \textbf{261.1M} & \textbf{84.4h} \\
\bottomrule
\end{tabular*}
\end{table}

\section{Perturbation Stress-Test Protocol Details}
\label{app:perturbation}

The variance decomposition in Sec.~\ref{sec:controlled_factorial} establishes that deliberate harness variation produces large HV. A complementary future test of the control-theoretic framing is whether agents exhibit the characteristic signature of closed-loop dynamical systems: disproportionate output instability in response to small controller-input perturbations. The following protocol follows directly from the theory; it has not been run, and we record it here for future work rather than as evidence for the thesis.

\textbf{Setup.} The planned experiment fixes GPT-5.4 and $H_2$ (the middle configuration from the variance decomposition experiment) and applies two classes of perturbation to the controller inputs. The first is context ordering: at each step, the top-$k$ retrieved context chunks returned by the context construction function $c$ are randomly shuffled before being passed to the model, so that the chunks themselves are unchanged and only their order varies. The second is tool output noise: numerical values in tool outputs are perturbed by a small multiplicative factor drawn from $\mathcal{N}(1, \epsilon^2)$ for $\epsilon \in \{0.01, 0.05, 0.10\}$, while text outputs and error codes are left unchanged. For each perturbation class, the protocol would run 50 independent trials per task on a 50-task subset of SWE-bench Verified, comparing them with 50 clean-baseline runs on the same tasks.

\textbf{Metrics.} We define the output instability score as the fraction of task trajectories whose final outputs differ from the clean-baseline outputs on that task, normalized by the perturbation magnitude where applicable. An agent that is robust to its controller inputs will produce instability scores close to the clean-baseline variance floor; an agent that is sensitive, as predicted by the dynamical-systems view, will produce instability scores that scale disproportionately with perturbation magnitude.

\textbf{Expected diagnostic value.} This stress test would not be another benchmark score. It would measure whether the controller has enough damping to absorb small changes in context ordering and tool observations. If $H_3$ shows lower output instability than $H_1$ under matched perturbations, that would provide direct evidence that verification, drift checks, and rollback mechanisms improve closed-loop robustness rather than merely shifting the average pass rate.

\section{Industry Evidence Summary}
\label{app:industry}

The pattern argued for in Sec.~\ref{sec:evidence_leaderboards} is even more visible in industry, where agent products compete on harness quality rather than model choice.

The managed agents architecture decouples the ``Brain'' (model and harness logic), ``Hands'' (execution environments), and ``Session'' (event log) into independent layers, reporting that harness assumptions go stale as models improve and that this decoupling reduced time-to-first-token by 60\% at p50 and over 90\% at p95~\cite{martinScalingManagedAgents2026}. These are infrastructure improvements rather than model improvements.

The APEX-Agents benchmark, which evaluates agents on 480 real professional tasks across banking, consulting, and law, found that agent failures were predominantly execution and orchestration problems rather than knowledge failures~\cite{vidgen2026apex}. Zero-score rates ranged from 40\% to 62\% across configurations using the same underlying model. The bottleneck was not what the model knew but how the harness managed execution.

The same pattern appears across public agent systems: SWE-agent exposes an agent-computer interface for repository navigation and editing~\cite{yang2024sweagent}, OpenHands packages software-development agents as an execution platform~\cite{wang2024openhands}, and Codex CLI and Claude Code wrap frontier models in product-specific runtime layers~\cite{openaiCodexCLI2025,anthropicClaudecode2025}; other public coding-agent systems including OpenCode~\cite{anomalyOpencodeOpenSource2025}, Terminus~\cite{harborTerminus22026}, and the Hermes Agent~\cite{HermesAgentAgent} make the same architectural choice.\footnote{An open-source personal-AI variant is OpenClaw, described at \url{https://openclaw.ai/}.} These systems draw from a small set of frontier model providers~\cite{DeepSeek_V4pdfDeepseekaiDeepSeekV4Pro2026,teamQwen36PlusRealWorld2026}, yet they differ sharply in context handling, tool boundaries, execution isolation, and recovery behavior. The differentiating factor is the harness layer.

\section{Mechanism Analysis from Trajectory Logs}
\label{app:mechanism}

The factorial experiment in Sec.~\ref{sec:controlled_factorial} reports aggregate variance and ranking statistics. This appendix records the trajectory-level mechanism analysis that supports the qualitative conclusions summarised in the main body.

A subtle pattern in Table~\ref{tab:factorial} also informs the leaderboard argument. Cross-model variance is smallest under $H_2$ ($1.17$~pp$^2$) and largest under $H_3$ ($4.22$~pp$^2$), with $H_1$ in between ($1.72$~pp$^2$). The $H_1$ vs $H_2$ comparison suggests that better-engineered closed-loop control can suppress model differences by absorbing variance through feedback; the $H_3$ result shows that some harness mechanisms (per-step verification, recovery) can also expose previously hidden model differences. This complicates the simple picture that better harnesses make model choice irrelevant: the relationship between controller quality and cross-model variance depends on which mechanisms the harness implements, which is itself part of why disclosure matters.

\textbf{Mechanism analysis.} The trajectory logs clarify the mechanism behind these numerical differences. The dominant $H_1 \to H_2$ pattern is reduced control noise rather than increased model knowledge: stricter JSON-only action formatting, smaller task-focused tool schemas, compressed recent-history context, and structured error feedback reduce malformed outputs, repeated exploration, and overly broad patches. In paired flip cases, $H_1$ often reaches a plausible local fix but fails because the trajectory exhausts its step budget, loses the relevant state in an append-only history, or introduces pass-to-pass regressions while trying to repair the target failure. $H_2$ more often keeps the model on a narrower action path: it retains recent task-relevant evidence, surfaces earlier relevant steps through retrieval, and makes tool / action boundaries clearer, producing smaller patches less likely to disturb retained tests. This is the qualitative counterpart of the variance result: $H_2$ does not make the model more knowledgeable, it makes the control loop better constrained.

The $H_2 \to H_3$ flips show a different mechanism. $H_3$ adds per-step verification, anomaly checks, and recovery signals, which turn false progress into visible feedback. $H_2$ sometimes accepts a patch that fixes the target test while breaking retained tests, or treats a failed command, empty test selection, or failed replacement as useful evidence. $H_3$ frequently spends more steps and tokens, but its verifier flags ineffective test commands, failed edits, incomplete fixes, and regression risks before submission. The additional $H_3$ gains therefore come primarily from closing the loop around verification and regression avoidance: the harness converts ambiguous execution traces into corrective observations that the same base model can act on.



\end{document}